\documentclass[conference]{IEEEtran}

\IEEEoverridecommandlockouts
\usepackage{cite}
\usepackage{amsmath,amssymb,amsfonts}
\usepackage{algorithmic}
\usepackage{graphicx}
\usepackage{textcomp}
\usepackage{xcolor}
\usepackage{hyperref}
\usepackage{cite}
\usepackage{amsmath,amssymb,amsfonts}
\usepackage{algorithmic}
\usepackage{graphicx}
\usepackage{textcomp}
\usepackage{multirow}
\usepackage{xcolor}
\def\BibTeX{{\rm B\kern-.05em{\sc i\kern-.025em b}\kern-.08em
    T\kern-.1667em\lower.7ex\hbox{E}\kern-.125emX}}

\begin{document}

\author{\IEEEauthorblockN{Sasho Nedelkoski\IEEEauthorrefmark{1},\
Jasmin Bogatinovski\IEEEauthorrefmark{1},
Alexander Acker\IEEEauthorrefmark{1},
Jorge Cardoso\IEEEauthorrefmark{2},
Odej Kao\IEEEauthorrefmark{1} 
}
\IEEEauthorblockA{\IEEEauthorrefmark{1}Distributed and Operating Systems, 
TU Berlin, Berlin, Germany\\ \{nedelkoski, jasmin.bogatinovski, alexander.acker, odej.kao\}@tu-berlin.de}
\IEEEauthorblockA{\IEEEauthorrefmark{2}Huawei Munich Research Center, Huawei Technologies, Munich, Germany \\ jorge.cardoso@huawei.com}
}

\title{Self-Attentive Classification-Based Anomaly Detection in Unstructured Logs}

\newcommand{\jasmin}[1]{{\color{green} Jasmin: #1}}
\maketitle

\begin{abstract}
The detection of anomalies is essential mining task for the security and reliability in computer systems.
Logs are a common and major data source for anomaly detection methods in almost every computer system. 
They collect a range of significant events describing the runtime system status. Recent studies have focused predominantly on one-class deep learning methods on predefined non-learnable numerical log representations. The main limitation is that these models are not able to learn log representations describing the semantic differences between normal and anomaly logs, leading to a poor generalization of unseen logs. We propose Logsy, a classification-based method to learn log representations in a way to distinguish between normal data from the system of interest and anomaly samples from auxiliary log datasets, easily accessible via the internet. The idea behind such an approach to anomaly detection is that the auxiliary dataset is sufficiently informative to enhance the representation of the normal data, yet diverse to regularize against overfitting and improve generalization.
We propose an attention-based encoder model with a new hyperspherical loss function. This enables learning compact log representations capturing the intrinsic differences between normal and anomaly logs. Empirically, we show an average improvement of 0.25 in the F1 score, compared to the previous methods.
To investigate the properties of Logsy, we perform additional experiments including evaluation of the effect of the auxiliary data size, the influence of expert knowledge, and the quality of the learned log representations. The results show that the learned representation boost the performance of the previous methods such as PCA with a relative improvement of 28.2\%.
\end{abstract}


\section{Introduction}
Anomaly detection~\cite{grubbs1969procedures, hodge2004survey, pimentel2014review} is a data mining task of finding observations in a corpus of data that differ from the expected behaviour. Anomalies in large systems such as cloud and high-performance computing (HPC) platforms can impact critical applications and a large number of users~\cite{zhang2015rapid}. Owing to the inevitable weaknesses in software and hardware, systems are prone to failures, which can potentially harm them to a large extent~\cite{chandola2009anomaly, sultana2018survey}. Timely and accurate detection of such threats is necessary for reliability, stable operation, and mitigation of losses in a complex computer system. 

Logs are an important data source for anomaly detection in computer systems~\cite{zhu2019tools, du2017deeplog, zhang2019robust}. They represent interactions between data, files, services, or applications, and are typically utilized by developers, and data-driven methods to understand system behaviours and to detect, localize, and resolve problems that may arise. Log messages have free-form text structure written by the developers, which record a specific system event describing the runtime system status. 
Specifically, a log message is a composition of constant string template and variable values originating from logging instruction (e.g., print("total of \%i errors detected", 5)) within the source code. 


A common approach for log anomaly detection is one-class classification~\cite{moya1993one}, where the objective is to learn a model that describes the normal system behaviour, usually assuming that most of the unlabeled training data is non-anomalous and that anomalies are samples that lie outside of the learned decision boundary. The massive log data volumes in large systems have renewed the interest in the development of one-class deep learning methods to extract general patterns from non-anomalous samples. Previous studies have been focused mostly on the application of long short-term memory (LSTM)-based models~~\cite{du2017deeplog, meng2019loganomaly, zhang2019robust}. They leverage log parsing~\cite{nedelkoski2020selfsupervised, he2017drain} on the normal log messages and transform them into log templates, which are then utilized to train the models. The formulated task is to predict the next index of the log template in the sequence $t_{m+1}$ by utilizing the history of template indices $H=t_0,\dots,t_m$. In other disciplines, numerous deep learning methods increase their performances by incorporating large amounts of data available through the internet. A common approach to use these data is unsupervised learning. In natural language processing (NLP), word2vec~\cite{mikolov2013distributed} and more recent language models BERT~\cite{devlin2018bert} are standard and responsible for significant improvements in various NLP tasks. These models are pretrained on large corpora of text such as Wikipedia and later fine-tuned on the particular task or dataset. Recent studies in log anomaly detection~\cite{meng2019loganomaly, zhang2019robust} utilize a pre-trained word embeddings to numerically represent the log templates instead of the integer log sequences~\cite{du2017deeplog}, where they observe small improvements in the prediction of unseen logs. 

However, the learning of the sequence of template indices and the enhanced log message embedding approaches still have large limitations in terms of generalization for previously unseen log messages. They tend to produce false predictions owing to the imperfect log vector representations. For example, learning sequence of indices fails to correctly classify a newly appearing log messages, and, the domain where the word vectors are pre-trained (e.g., Wikipedia) has essential differences from the language used in computer system development. To partly mitigate some of these limitations, a possibility is to incorporate labeled data from operators and perform life-long learning~\cite{du2019lifelong}. Yet, it still requires frequent periodical retraining, updates, and costly expert knowledge to label the data, without addressing the problem of generalization on unseen logs that appear between retraining epochs.

Often, the assumption for the normal data in anomaly detection methods is that it should be compact~\cite{ruff2019deep}. This means the normal log messages should have vector representations with close distances between each other, e.g., concentrated within a tight sphere, and the anomalies should be spread far from the distribution of the normal samples. We propose a new anomaly detection method that directly addresses the challenge of obtaining representative and compact numerical log embeddings. We train a neural network to learn log vector representations in a manner to separate the normal log data from the system of interest and log messages from auxiliary log datasets from other systems, easily accessible via the internet. The concept of such a classification approach to anomaly detection is that the auxiliary dataset helps learn a better representation of the normal data while regularizing against overfitting. This ultimately leads to a better generalization in unseen logs. For example, for a target system logs of interest $T$ where anomaly detection needs to be performed, as auxiliary data could be employed one or more datasets from an open-source log repository (e.g.,~\cite{oliner2007supercomputers}). As a neural network architecture, we adopt the Transformer encoder with multi-head self-attention mechanism~\cite{vaswani2017attention}, which learns context information from the log message in the form of log vector representations (embeddings). We propose a hyperspherical learning objective that enforces the model to learn compact log vector representations of the normal log messages. This enforces for the normal samples to have concentrated (compact) vector representations around the centre of a hypersphere. It enables better separation between the normal and the anomaly data, where a distance from the centre of such a sphere is used to represent an anomaly score. Small distances correspond to normal samples, while large distances correspond to anomalies. The method also enables a direct log-to-vector transformation, which can be used to improve the performances of previous related methods. Additionally, it allows the operator to intervene and correct misclassified samples, which could be used for the next retraining of the model. 

The contributions of this study can be summarized in the following points. 
\begin{enumerate}
    \item A new classification-based method for log anomaly detection utilizing self-attention and auxiliary easy-accessible data to improve log vector representation.
    \item Modified objective function using hyperspherical decision boundary, which enables compact data representations and distance-based anomaly score.
    \item The proposed approach is evaluated against three real anomaly detection datasets from HPC systems, Blue Gene/L, Thunderbird, and Spirit. The method significantly improves the evaluation scores compared to those in the previous studies. 
    \item In another set of experiments, an investigation of the effects of variations in the amount of auxiliary data for anomaly detection and inclusion of labelled data is performed.
    \item We provide an open-source implementation of the method.
\end{enumerate}

\section{Related Work}\label{relatedwork}
A significant amount of research and development of methods for log anomaly detection has been published in both industry and academia~\cite{liang2007failure, du2017deeplog, xu2009detecting,  zhang2019robust, meng2019loganomaly, nedelkoski2020selfsupervised}.
Supervised methods were applied in the past to address the log anomaly detection problem. For example, ~\cite{liang2007failure} applied a support vector machine (SVM) to detect failures, where both normal and anomalous samples are assumed to be available. For an overview of supervised approaches to log anomaly detection we refer to Brier et al.~\cite{breier2015anomaly}. However, obtaining system-specific labelled samples is costly and often practically infeasible. 

Several unsupervised learning methods have been proposed as well. Xu et al.~\cite{xu2009detecting} proposed using the Principal Component Analysis (PCA) method, where they assume that there are different sessions in a log file that can be easily identified by a session-id attached to each log entry. It first groups log keys by session and then counts the number of appearances of each log key value inside each session. A session vector is of size $n$, representing the number of appearances for each log key in $K$ in that session. A matrix is formed where each column is a log key, and each row is one session vector. PCA detects an abnormal vector (a session) by measuring the projection length on the residual subspace of a transformed coordinate system. The publicly available implementation allows for the term frequency-inverse document frequency (TF-IDF) representation of the log messages, utilized in our experiments as a baseline.  Lou et al.~\cite{lou2010mining} proposed Invariant Mining (IM) to mine the linear relationships among log events from log event count vectors. 

The wide adoption of deep learning methods resulted in various new solutions for log-based anomaly detection. Zhang et al.~\cite{Zhang2016AutomatedIS} used LSTM to predict the anomaly of log sequence based on log keys. Similar to that, DeepLog~\cite{du2017deeplog} also use LSTM to forecast the next log event and then compare it with the current ground truth to detect anomalies. Vinayakumar et al.~\cite{Vinayakumar2017LongSM} trained a stacked-LSTM to model the operation log samples of normal and anomalous events. However, the input to the unsupervised methods is a one-hot vector of logs representing the indices of the log templates. Therefore, it cannot cope with newly appearing log events. 

Some studies have leveraged NLP techniques to analyze log data based on the idea that log is a natural language sequence. Zhang et al.~\cite{Zhang2016AutomatedIS} proposed to use the LSTM model and TF-IDF weight to predict the anomalous log messages. Bertero et al.~\cite{bertero2017experience} used word2vec and traditional classifiers, like SVM and Random Forest, to check whether a log event is an anomaly or not. Similarly, LogRobust~\cite{zhang2019robust} and LogAnomaly~\cite{meng2019loganomaly} incorporate pre-trained word vectors for learning of a sequence of logs where they train an attention-based Bi-LSTM model.


Different from all the above methods, we add domain bias on the anomalous distribution to improve detection~\cite{steinwart2005classification}. We provide such bias by employing easily accessible log datasets as an auxiliary data source. We evaluate Logsy against unsupervised approaches, as even it is a classification based approach, it does not use labels from the target system, which as mentioned are often infeasible to obtain. From the perspective of using labels of the target system it is an unsupervised approach.

\section{Towards Classification-Based Log Anomaly Detection}
Anomaly detection can be also viewed as density level set estimation~\cite{tsybakov1997nonparametric}. Steinwart et al.~\cite{steinwart2005classification} state that this can be interpreted as binary classification between the normal and the anomalous distribution and point out that the bias on the anomalous distribution is essential for improved detection. Meaning that if we provide some information to the model of how anomalous data looks like, it will boost its performance. For instance, we may interpret the class assumption that semi-supervised anomaly detection approaches require on the anomalies, as such prior knowledge~\cite{ruff2019deep}. Moreover, specific types of data can have an inherent properties that allows us to make more informed prior assumptions such as the word representations in texts~\cite{bengio2013representation}. Here the assumption is that each word meaning depends on its context.

We assume that drawing realistic samples from some auxiliary easy-access corpus of log data, can be much more informative for an added description of normal and anomalies compared to sampling noise, or no data used. The use of auxiliary data adds extra value to the method, while preserving the information from the normal data.

\textsc{\textbf{Problem Definition}}. \textit{Let $\mathcal{D}=\{(\mathbf{x_1}, y_1), \dots, (\mathbf{x_n}, y_n)\}$ be the training logs from the system of interest where $\mathbf{x_i} \in \mathbb{R}^d$ is a log message where it words are represented in $d-dimensional$ space (the log message is represented by $d\times \vert r \vert$ matrix, where $\vert r \vert$ is number of words) and $y_i=0; 1 < i \leq n$, assuming that the data in the system of interest is mostly composed of normal samples. Let $\mathcal{A}=\{(\mathbf{x_n}, y_{n}),\dots, (\mathbf{x_{n+m}}, y_{n+m})\}$, where $m$ is the size of the auxiliary data and $y_i={1}; n < i \leq n+m$. Let $\phi(\mathbf{x_i}, y_i, \theta): \mathbb{R}^d \rightarrow \mathbb{R}^p$ be a function represented by a neural network, which maps the input log message embeddings to vector representations in $\mathbb{R}^p$, and $l: \mathbb{R}^p \rightarrow [0, a], a \in \mathbb{R}$ be a function, which maps the output to an anomaly score. The task is to learn the parameters $\theta$ from the training data, and then for each incoming instance in the prediction phase $\mathcal{D}_t=\{(\mathbf{x_1^t}), (\mathbf{x_2^t}),\dots, (\mathbf{x_j^t}), \dots\}$, $t$ indicates test sample, predict whether it is anomaly or normal based on the anomaly scores obtained by $l(\phi(\mathbf{x_i}, y_i, \theta))$.}

\section{Self-Attentive Anomaly Detection with Classification-based Objective}
In this section, we explain the proposed method in detail. We provide formal definitions needed for explaining the method. We describe the data preprocessing, the neural network, the log vector representations, and how they are utilized in the modified objective function for anomaly detection.

\subsection{Preliminaries}\label{preliminaries}
We define a log as a sequence of temporally ordered unstructured text messages $L=(x_{i} \,:\,i=1,2,...)$, where each message $x_{i}$ is generated by a logging instruction (e.g. printf(), log.info()) within the software source code, and $i$ is its positional index within the sequence. The log messages consist of a constant and an optional varying part, respectively referred to as log template and variables.

The smallest inseparable singleton object within a log message is a token. Each log message consists of a finite sequence of tokens, $\mathbf{r_i}=(w_{j}\,:\,w_{j} \in  \mathbb{V},\,j=1,2,..., s_i)$, where $ \mathbb{V}$ is a set (vocabulary) of all tokens, $j$ is the positional index of a token within the log message $x_i$, and $s_i$ is the total number of tokens in $x_i$. We use $\vert r_i \vert$ instead of $s_i$ in following. For different $x_i$, $\vert \mathbf{r_i} \vert$ can vary. Depending on the concrete tokenization method, $w_j$ can be a word, word piece, or character. Therefore, tokenization is defined as a transformation function $\mathcal{T}: x \to \mathbf{r}$.

With respect to our proposed method, the notions of context and numerical vector representation (embedding vector) are additionally introduced. Given a token $w_j$, its context is defined by a preceding and subsequent sequence of tokens, i.e. a tuple of sequences: $C(w_j)=((w_{1}, w_{2},...,w_{j-1}), \allowbreak (w_{j+1}, w_{j+2},...,w_{\vert \mathbf{r_i} \vert}))$, where $0 \leq j \leq \vert \mathbf{r_i} \vert$. An embedding vector is a $d$-dimensional real valued vector representation $\mathbf{s} \in \mathbb{R}^{d}$ of either a token or a log message.

In the learned vector space, similar log messages should be represented by closer embedding vectors while largely different log messages should be distant. For example, the embedding vectors for "Took 10 seconds to create a VM" and "Took 9 seconds to create a VM" should have a small distance in $d$-dimensional space, while vectors for "Took 9 seconds to create a VM" and "Failed to create VM 3" should be distant.

We refer to the data from the system of interest as target dataset, i.e., the system where we want to detect anomalies. Important to note is that we are not using any anomaly data from the target system for learning purposes in our experiments. The term auxiliary data refers to other non-related systems, which serve only for training the model. All the results during test time are performed on a test set extracted from the target dataset.

\subsection{Logsy}
\begin{figure}[htbp]
\centerline{\includegraphics[width=\columnwidth]{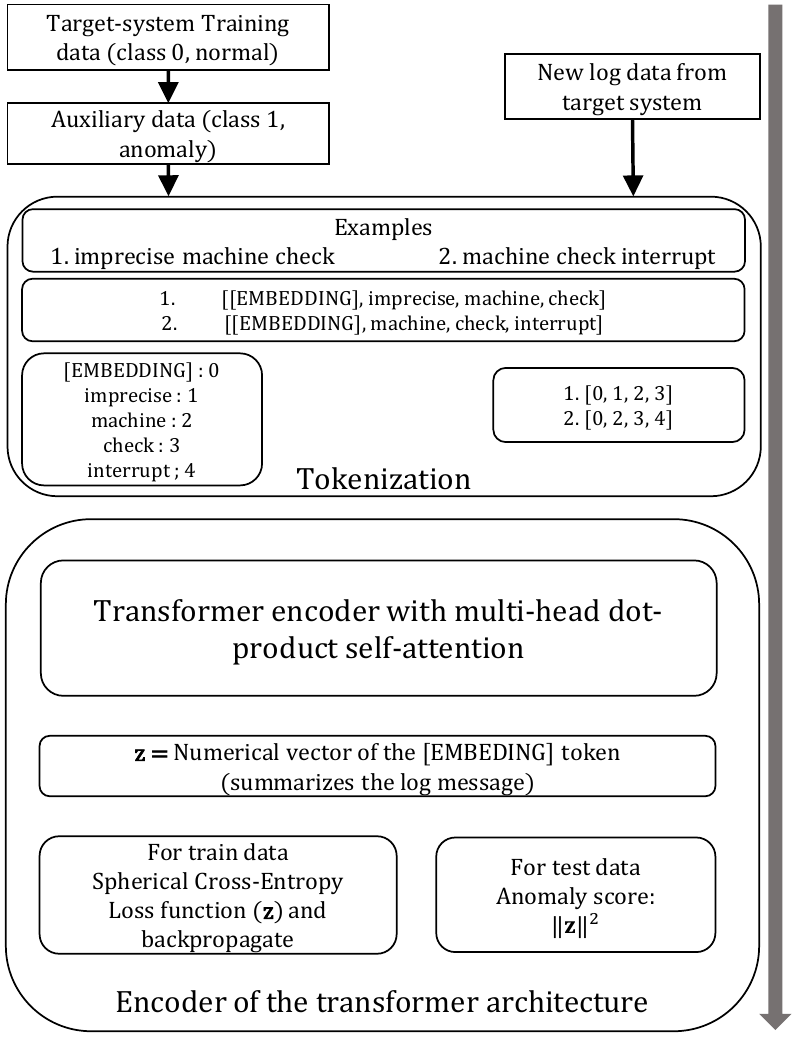}} 
\caption{Overview of the architecture and component details of Logsy.}
\label{overviewmethod}
\end{figure}

The method is composed of two main parts, the tokenization of the log messages and the neural network model. In the following section, we discuss the inner workings of the proposed method, which is depicted in~\figurename~\ref{overviewmethod}. 

\textbf{Tokenization.} Tokenization transforms the raw log messages into a sequence of tokens, as shown in~\figurename~\ref{overviewmethod}. For this purpose, we utilize the standard text preprocessing library NLTK~\cite{loper2002nltk}. The message is first filtered for HTTP and system path endpoints (e.g., /p/gb2/stella/RAPTOR/). Every capital letter is converted to a lower letter, and all of the ASCII special characters are removed. The log message is split into word tokens. We remove every token that contains numerical characters, as they often represent variables in the log message and are not informative. Additionally, we remove the most commonly used English words that are in the stop words dictionary of NLTK (e.g., \textit{the} and \textit{is}). To the front of the tokenized log message, a special '[EMBEDDING]' token is added. In the model, the '[EMBEDDING]' token attends overall original tokens from the sample, which enables the model to summarize the context of the log message in the vector representation. All tokens from every log message form vocabulary $\mathbb{V}$ of size $|\mathbb{V}|$, where each token is represented with integer label $i \in {0, 1, \dots,|\mathbb{V}|-1}$. An important advantage of Logsy compared to previous approaches is that it does not depend on log parsers as a pre-processing step. We consider the tokenized log message as direct input to the model. The advantage is that there is no loss of information from the log message, due to the imperfections that exist in the log parsing methods.

\textbf{Model}. Logsy has two operation modes -- offline and online. During the offline phase, log messages are used to tune all model parameters via backpropagation and optimal hyper-parameters are selected. During the online phase, every log message is passed forward through the saved model. This generates the respective log vector representation $\mathbf{z}$ and an anomaly score for each message.

\begin{figure}[!t]
\centerline{\includegraphics[width=0.9\columnwidth]{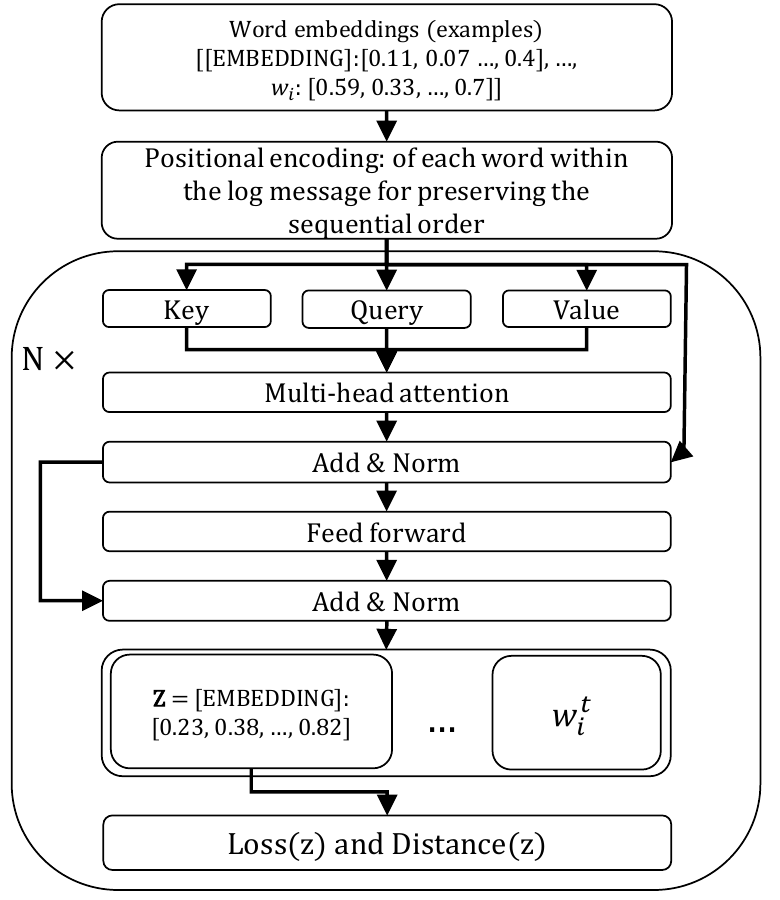}} 
\caption{Transformer encoder architecture with multi-head self-attention.}
\label{multiheadattention}
\end{figure}

As depicted in \figurename~\ref{multiheadattention}, the model applies two operations on the input tokens: token vectorization (word embeddings) and positional encoding. The subsequent structure is the encoder of the Transformer~\cite{vaswani2017attention} module with multi-head self-attention, which takes the result of these operations as input. 
At the output of the encoder, there are $\vert r_i \vert$ transformed vector representation from the initial tokens. Recall that the '[EMBEDDING]' token has its transformed representation, which is used as a final log vector representation. We denote the size of this vector as $d$. This also represents the size of all the layers of the model and the word embeddings. The last two parts are the objective (loss) function during training and the computation of the anomaly score for test-time samples. Based on the loss, gradients are back-propagated to tune the parameters of the model, while based on the anomaly score we decide if the sample is anomalous or normal. In the following, we provide a detailed explanation of each element of the method. \figurename~\ref{multiheadattention} depicts the inner working of the transformer encoder.

Since all subsequent elements of the model expect numerical inputs, we initially transform the tokens into randomly initialized numerical vectors $\mathbf{x} \in \mathbb{R}^d$. These vectors are referred to as token embeddings and are part of the training process, which means they are adjusted during training to represent the semantic meaning of tokens depending on their context. These numerical token embeddings are passed to the positional encoding block. In contrast to e.g., recurrent architectures,  attention-based models do not contain any notion of input order. Therefore, this information needs to be explicitly encoded and merged with the input vectors to take their position within the log message into account. This block calculates a vector $\mathbf{n} \in \mathbb{R}^d$ representing the relative position of a token based on a sine and cosine function.

\begin{equation}
    n_{2k}=sin \left( \frac{j}{10000^{\frac{2k}{d}}} \right), \;\;
    n_{2k+1}=cos \left( \frac{j}{10000^{\frac{2k + 1}{d}}} \right).
\end{equation}

Here, $k=0,1,\dots,d-1$ is the index of each element in $\mathbf{n}$ and $j=1,2,\dots, |r_i|$ is the positional index of each token. Within the equations, the parameter $k$ describes an exponential relationship between each value of vector $\mathbf{n}$. 
The applied sine and cosine functions allow for better discrimination of the respective values within a specific vector of $\mathbf{n}$. They have an approximately linear dependence on the position parameter $j$, which is hypothesized to make it easy for the model to attend to the respective positions. Finally, both vectors can be combined as $\mathbf{x'} = \mathbf{x} + \mathbf{n}$.  We summarize all token embedding vectors of a log message as matrix rows $\mathbf{x'}^T \in X'$ on which the following formula is applied:

\begin{equation}
    X''_l=softmax \left( \frac{Q_l \times K^T_l}{\sqrt{w}} \right) \times V_l, \; \text{for} \; l = 1, 2, \dots, L.
\end{equation}
Thereby, $L$ denotes the number of attention heads, $w = \frac{d}{L}$ and $d \, \text{mod} \, L = 0$. The parameters $Q$, $K$ and $V$ are matrices, that correspond to the query, key, and value elements in \figurename~~\ref{multiheadattention}. They are obtained by applying matrix multiplications between the input $X'$ and respective learnable weight matrices $W_{l}^{Q}$, $W_{l}^{K}$, $W_{l}^{V}$:

\begin{equation}
    Q_l= X' \times W_{l}^{Q}, \; K_l= X' \times W_{l}^{K}, \; V_l= X' \times W_{l}^{V},
\end{equation}

where $W_{l}^{Q}, \; W_{l}^{K}, \; W_{l}^{V} \in \mathbb{R}^{M \times w}$. The division by $\sqrt{w}$ stabilizes the gradients during training. After that, the softmax function is applied and the result is used to scale each token embedding vector $V_{l}$. The scaled matrices $X''_l$ are concatenated to a single matrix $X''$ of size $M \times d$. 

As depicted in \figurename~\ref{multiheadattention} there is a residual connection between the input token matrix $X'$ and its respective attention transformation $X''$, followed by a normalization layer $norm$. These are used for improving the performance of the model by tackling different potential problems encountered during the learning such as small gradients and the covariate shift phenomena. Based on this, the original input is updated by the attention-transformed equivalent as $X' = norm(X' + X'')$.

The last element of the encoder consists of two feed-forward linear layers with a ReLU activation in between. It is applied individually on each row of $X'$. Thereby, identical weights for every row are used, which can be described as a convolution over each attention-transformed matrix row with kernel size one. This step serves as additional information enrichment for the embeddings. Again, a residual connection followed by a normalization layer between the input matrix and the output of both layers is employed. This model element preserves the dimensionality $X'$.

The final element of the model consists of a single linear layer. It receives the encoder result $X'$ and extracts the token embedding vector of the ['EMBEDDING']. Since every log message token sequence is pre-padded by this special token, it is the first row of the matrix, i.e. $\mathbf{x'_{i, 0}} \in X', \forall i$. This vectors are the log vector representations and are used in the objective function and as well as log message embeddings.

\subsection{Objective function}
To ensure learning of the intrinsic differences of normal and anomaly log samples, we propose a spherical loss function. It is designed to integrate the previously mentioned assumption that normal data is often concentrated having close distances between the normal samples, while also learning properties to distinct from anomalous samples. This is done by employing a radial classification loss which enforces a compact hyperspherical decision region for the normal samples. 

To derive the loss, we start with the standard binary cross entropy. Let $\mathcal{D}=\{(\mathbf{x_1}, y_1), \dots, (\mathbf{x_{n+m}}, y_{n+m})\}$ be the concatenation of the training logs from the system of interest and the auxiliary data with $\mathbf{x_i} \in \mathbb{R}^{d\times|r_i|}$, where $|r_i|$ is the number of tokens in the log message and each token is a vector represented in $d-dimensional$ space. $y_i \in \{0,1\}$, and $y_i=0$ denotes normal samples (target system), while $y_i=1$ denotes an anomaly (auxiliary data). Let $\phi(\mathbf{x_i}, \theta): \mathbb{R}^d \rightarrow \mathbb{R}^p$ be our encoder architecture that maps the $|x_i|$ word embeddings form the log message to $p-dimensional$ vector. Let $l:\mathbb{R}^p \rightarrow [0,1]$ be a function which maps the output to an anomaly score. Using $\phi(\mathbf{x_i}, \theta)$ and $l(\cdot)$,d the standard binary cross-entropy loss can be written as:

\begin{equation}
    - \frac{1}{n}\sum_{i=1}^{n}(1-y_i)\log l(\phi(\mathbf{x_i}; \theta)) + y_i\log (1-l(\phi(\mathbf{x_i}; \theta)))
\end{equation}\label{bce}

For standard classifier function the $p-dimensional$ representation is transformed via linear layer followed by sigmoid activation function:

\begin{equation}\begin{aligned}
    - \frac{1}{n}\sum_{i=1}^{n}(1-y_i)\log((1+\exp(-\mathbf{w}^T \phi(\mathbf{x_i}, \theta)))^{-1}) \\
    +  y_i\log(1-(1+\exp(-\mathbf{w}^T \phi(\mathbf{x_i}, \theta)))^{-1})\end{aligned}
\end{equation}\label{bcesigmoid}

In the standard binary classifier with sigmoid function, the decision boundary is half-space.
The representation of the log messages is not guaranteed to be compact in this case. It could be very possible that the normal samples are scattered through the space with varying, potentially very large distances between them. To enforce compactness of the representations of the log messages we utilize the Gaussian radial basis function as $l(\cdot)$:

\begin{equation}
    l(\mathbf{z})= \exp(-\Vert \mathbf{z} \Vert ^2)
\end{equation}

Replacing the function into the loss function we get the hyper-spherical classifier:
\begin{equation}\begin{aligned}
    \frac{1}{n}\sum_{i=1}^{n}(1-y_i)\Vert \phi(\mathbf{x_i}; \theta) \Vert ^2 \\
    -  y_i\log(1-\exp(-\Vert \phi(\mathbf{x_i}; \theta) \Vert ^2))\end{aligned}
\label{hce}
\end{equation}

This ensures compactness of the normal samples, which will be enforced to be around the center of a sphere $\mathbf{c}=\mathbf{0}$. For normal samples, i.e., $y_i=0$, the loss function will minimize the distance to $\mathbf{c}$. This results in low values for the left term in Equation~\ref{hce}. In contrast, the right term of the loss function favors large distances for the anomalous samples. 
The center of a sphere $c$ could be any constant value, which is not relevant during the optimization.



\begin{table*}[htbp]
\centering
\caption{Dataset details.}
\label{datasets}
\begin{tabular}{c|c|c|c|c|c|c|c|c|c}
\hline
\multirow{2}{*}{\textbf{System}} & \multirow{2}{*}{\textbf{\#Messages}} & \multirow{2}{*}{\textbf{\#Anomalies}} & \multirow{2}{*}{\textbf{\#Anomalies5m}} & \multicolumn{5}{c|}{\begin{tabular}[c]{@{}c@{}}\textbf{\#Unique  Log  messages} \\ \textbf{in test  and  not  in  train  for  every  split}\end{tabular}} & \multirow{2}{*}{\begin{tabular}[c]{@{}c@{}}\textbf{total unique}\\ \textbf{messages}\end{tabular}} \\ \cline{5-9}
                        &                             &                              &                                & 10\%                      & 20\%                      & 40\%                      & 60\%                     & 80\%                     &                                                                                  \\ \hline
Blue Gene/L             & 4747963                     & 348460                       & 348460                         & 2679                      & 2621                      & 2256                      & 2231                     & 465                      & 4486                                                                             \\ 
Thunderbird             & 211212192                   & 3248239                      & 226287                         & 334                       & 127                       & 71                        & 27                       & 12                       & 3279                                                                             \\
Spirit                  & 272298969                   & 172816564                    & 764890                         & 1091                      & 1028                      & 297                       & 129                      & 73                       & 3441                                                                             \\ \hline
\end{tabular}

\end{table*}

A possible problem that usually arises in such spherical classifiers~\cite{ruff2019deep} is that the model is prone to learn trivial solutions by mapping the inputs to output a constant vector, i.e. $c$. However, the proposed loss function will not find the trivial solution because of the second term in the equation, representing the auxiliary data or the anomalies. To formally show that, let $\phi(\cdot)$ be the encoder network, which maps every log message to $\mathbf{c}$. It follows that that $\phi(\cdot)=\mathbf{0}$. In this case, the second term in Equation~\ref{hce} for $y_i=1$ will be infinity in the limit, which acts as a regularizer and prevents learning $\mathbf{c}$ as a trivial vector representation.


\subsection{Anomaly score and detecting anomalies}
Considering that the assumption of the objective function enforces compact, close to the center of the sphere $\mathbf{c}=\mathbf{0}$, representations, we define our anomaly score as the distance of the log vectors (obtained from the 'EMBEDDING' token) to the center $\textbf{c}$ of the hypersphere.

\begin{equation}
    A(\mathbf{x_i}) = \Vert \phi(\mathbf{x_i}; \theta) \Vert ^2
\end{equation}

We define low anomaly scores $A(\mathbf{x_i})$ to be normal log messages, while large scores stand for the anomalies. To decide if the sample is anomalous or normal, we use a threshold $\mathcal{E}$. If the anomaly scores $A(\mathbf{x_i}) > \mathcal{E}$, then the sample is an anomaly, otherwise, we consider it as normal. This concludes the explanation of the inner workings of the method. In the following, we describe two properties of the model.

\subsection{Including expert knowledge}
Most computer systems, are to some extend, supervised and operated by an administrator. Over time, the administrator can manually inspect a small portion of the log events and provide labels.
As additional option, Logsy allows incorporation of such labels from the target system. The second term in Equation~\ref{hce}, used for the auxiliary data, could be also utilized for the inclusion of operator-labeled samples. This enables the addition of even more realistic, however, costly anomaly samples that help to learn the anomaly distribution, and further improve the performance. The labeled samples either need to be added together with the auxiliary data and retrain, or pre-training the model with the normal and auxiliary data followed by fine-tuning with the labeled data. With such a training procedure, the model extracts the relevant information from the auxiliary data and already learns good log representations for anomaly detection, as later shown in the experiments. The replacement of the auxiliary data with the labeled samples allows the model to only fine-tune its parameters in a few epochs. This preserves the already learned information from the larger auxiliary dataset as a bias to the fine-tuning procedure. In the experiments, we show that the inclusion of a small portion of labeled samples improves the performance of the model.




\subsection{Vector representations of the logs}\label{vectorrepresentations}
Learning numerical vector representations from the logs is fundamental for the performance of any machine learning method for log anomaly detection. Logsy can be utilized for obtaining such numerical log representations. These representations are used by the objective function of the method, to perform anomaly detection, but could be as well used to replace other, less powerful representations (e.g., TF-IDF in previous log-based anomaly detection methods such as the PCA ~\cite{xu2009detecting}), aiming to enhance their anomaly detection. 


\begin{figure}[htbp]
    \centering
    \includegraphics[width=0.6\columnwidth]{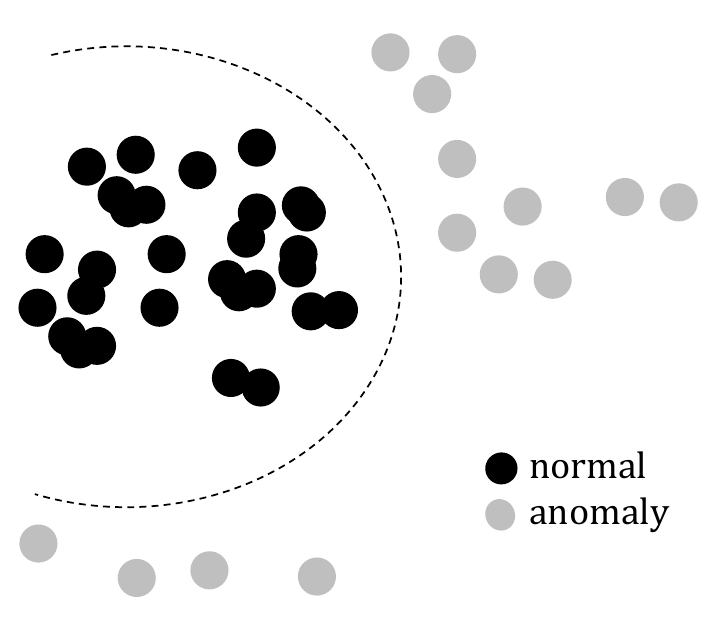}
    \caption{Ideal distribution of the log vector representations in space.}
    \label{fig:lowdimensions}
\end{figure}

The transformed vector of the ['EMBEDDING'] token is used for representing the context of the log message, which is the only output of the model to the loss function. Thus, it is forced to summarize the log message. By using the spherical classification decision boundary, we enforce the normal samples to be close to each one and compactly represented around the center of the sphere. This leaves the anomalies to disperse around the spherical decision boundary in the high-dimensional space. In \figurename~\ref{fig:lowdimensions}, we illustrate a lower-dimensional plot of how the ideal log representations should look like. A decision boundary (dashed line) can be drawn to optimally separate the classes. We demonstrate such behavior in the evaluation section on real data with Logsy, where we show how the normal and abnormal samples are distributed in low dimensional space.

\section{Evaluation}
To quantify the performance of Logsy, we perform a variety of experiments. We compare the method against two publicly available baselines DeepLog and PCA on three real-world HPC log datasets. We describe the main properties of the datasets, discuss the experimental setup, and present the results. We empirically and qualitatively evaluate the log vector representations from Logsy, where we utilize them in the PCA method and observed improved performance. Logsy is evaluated against unsupervised approaches, as from the perspective of using labels of the target system, it is an unsupervised approach.

\subsection{Experimental setup}
We select three open real-world datasets from HPC systems for evaluation as target systems, namely Blue Gene/L, Spirit, and Thunderbird~\cite{oliner2007supercomputers}. They share an important characteristic associated with the appearance of many new log messages in the timeline of the data, i.e., the systems change over time. Furthermore, as an additional dataset for enriching the auxiliary data in all experiments we use the HPC RAS log dataset~\cite{zheng2011co}. Due to the absence of labels this dataset cannot be used for evaluation purposes--can not be a target dataset.

For each target dataset as an auxiliary data to represent the anomaly class we use logs from the remaining datasets. It is important to note that the target vs auxiliary splits, ensure that there is no leak of information from the target system into the auxiliary data. Meaning, there are no labeled samples from the target system into the auxiliary data. These logs consist only of easily accessible logs from other systems via the internet. 
The non-anomalous samples from the target system are the target dataset. For example, when Blue Gene/L is our system of interest (i.e., the target system) proportion of the negative samples of Thunderbird, Spirit, and RAS are used as an auxiliary dataset to represent the anomaly class. These auxiliary samples could be also error messages obtained from online code repositories (e.g., GitHub).
We perform anomaly detection on the test samples from the target dataset for determining the scores. 

The datasets are collected between 2004 and 2006 on three different supercomputing systems: Blue Gene/L, Thunderbird, and Spirit.  The logs contain anomaly and normal messages identified by anomaly category tags and are therefore amenable to anomaly detection and prediction research. All systems were ranked on the Top500 Supercomputers List at the time (as of June 2006). The various machines are produced by IBM, Dell, Cray, and HP. All systems were installed at Sandia National Labs (SNL), except Blue Gene/L, which is at Lawrence Livermore National Labs (LLNL). In Table~\ref{datasets} we summarize the main characteristics of the datasets.

\begin{figure*}[!t]
    \centering
    \includegraphics[width=1.0\textwidth]{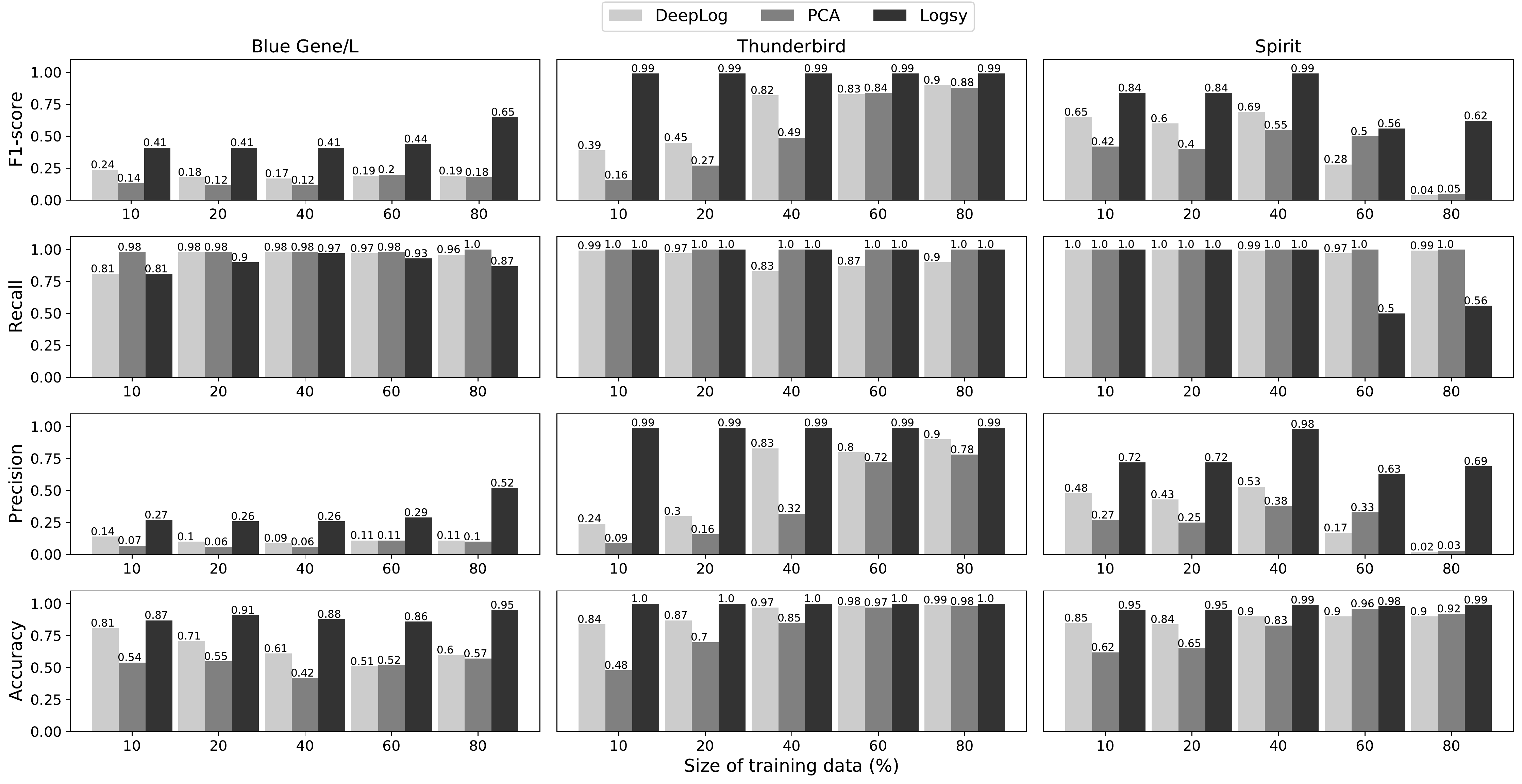}
    \caption{Comparison of the evaluation scores against the two baselines DeepLog and PCA on three different datasets.}
    \label{fig:results}
\end{figure*}

Table~\ref{datasets} shows that Thunderbird and Spirit are quite large datasets of more than 200 million log messages. For computation-time purposes we restrict the data size on the first 5 million, when sorted by timestamp, log messages. We ensure that the 5 million log lines preserve the properties of the dataset, as shown in Table~\ref{datasets}, which is that new unseen logs appear in the test data split. The Blue Gene/L dataset has less than 5 million messages, thus we keep it in total. \#Anomalies5m shows the number of anomalous log messages in those 5 million messages.


To evaluate the robustness and generalization of Logsy in detail, we conduct several experiments with different train-test splits on the target dataset. To ensure that the test data contains new log messages previously unseen in the training we always split the data when sorted by the timestamp of the log messages. We perform 5 different data splits to cover as many possible scenarios, i.e., the first 10\% training; 90\% test data, 20\% training -- 80\% test, 40\% training -- 60\% test, 60\% training -- 40\% test, and 80\% training -- 20\% test. 


The number of unique log messages after tokenization is presented in Table~\ref{datasets}. We observe that in every split there are new previously unseen log messages that appear in the test data, which is the main point for empirically proving generalization. Decreasing the size of the training data increases the number of novel log messages in the test split.

\subsubsection{Evaluation methods}
To enable comparability between our method to the previous work, we adopt the standard evaluation scores. We evaluate our method in F1-score, precision, recall, accuracy, which depends on the true negatives (TN), true positives (TP), false negatives (FN), and false positives (FP) predictions. The positive class of 1, is assumed to be an anomalous log. 

\subsubsection{Baselines}
We compare Logsy against two publicly available baseline methods, i.e., PCA~\cite{xu2009detecting} and Deeplog~\cite{du2017deeplog}. The current claimed state-of-the-art method LogAnomaly~\cite{meng2019loganomaly} to best of our knowledge has no publicly available implementation, as it is industry-related research. Moreover, LogAnomaly reports only marginal improvement over DeepLog of 0.03 F1 score, and thus both approaches are relatively comparable. The parameters of these methods are all tuned to produce their best F1 score.

\subsubsection{Logsy: Implementation details}
Every log message during tokenization is truncated to a maximum of $\max(\vert r_i \vert)=50$ tokens. Logsy has two layers of the transformer encoder, i.e., N=2 in \figurename~\ref{overviewmethod}. The words are embedded with 16 neural units, and the higher level vector representations obtained with the transformer encoding are all of the same sizes. The size of the feed-forward network that takes the output of the multi-head self-attention mechanism is also 16, which makes the '[EMBEDDING]' vector the same size. For the optimization procedure for every experiment, we use a dropout of 0.05, Adam optimizer with a learning rate of 0.0001, weight decay of 0.001. We address the imbalanced number of normal versus anomaly samples with adding weights to the loss function for the two classes, 0.5 for the normal and 1.0 for the anomaly class. The models are trained until convergence and later evaluated on the respective test split.

\subsection{Results and discussion}
We show the overall performance of Logsy compared to the baselines in \figurename~\ref{fig:results}. Generally, Logsy achieves the best scores, having an averaged F1 score in all the splits of 0.448
on the Blue Gene/L dataset, 0.99 on the Thunderbird dataset, and 0.77 on the Spirit data. Both DeepLog and PCA, have lower F1 scores in all experiments performed. It is shown that the baselines have a very high recall, but also low precision. This means they can find the anomalies, however, producing large amounts of false-positive predictions. 
Logsy, on the other hand, preserves the high recall across the datasets and evaluation scenarios but shows a large improvement in the precision scores. This is due to the correct classification of new unseen log messages and the reduction of the false positive rate. For instance, on the Blue Gene/L dataset, DeepLog and PCA respectively show 2-4 times lower precision compared to Logsy. Overall, Logsy is the most accurate method having an average of 0.9. If a log anomaly detection method generates too many false alarms, it will add too much overhead to the operators and a large amount of unnecessary work. Therefore, high precision methods are favourable. DeepLog leverage the indexes of log templates, which ignore the meaning of the words in the log messages, to learn the anomalous and normal patterns. However, different templates having different indexes can share common semantic information and both could be normal. Ignoring this information results in the generation of false positives for DeepLog compared to Logsy.

We notice that increasing the training size also increases the F1 score in almost all methods, except for the last two splits in Spirit. These splits are unfortunate as they have a very small number of anomalies. Important to note is that Logsy outperforms the baselines even when only 10\% of the data is training data. For example, in Blue Gene/L we have 0.32 F1-score on 10\% training data, while the largest F1-score of the baselines is 0.24. In Thunderbird, this difference is even more noticeable, where an F1-score of 0.99 is already achieved in with the first 10\%. This shows that even with a small amount of training data from the target system, Logsy extracts the needed information of what causes a log message to be normal or anomaly, and produces accurate predictions even in unseen samples.

\begin{figure*}[htbp]
\minipage{0.33\textwidth}
  \includegraphics[width=\linewidth]{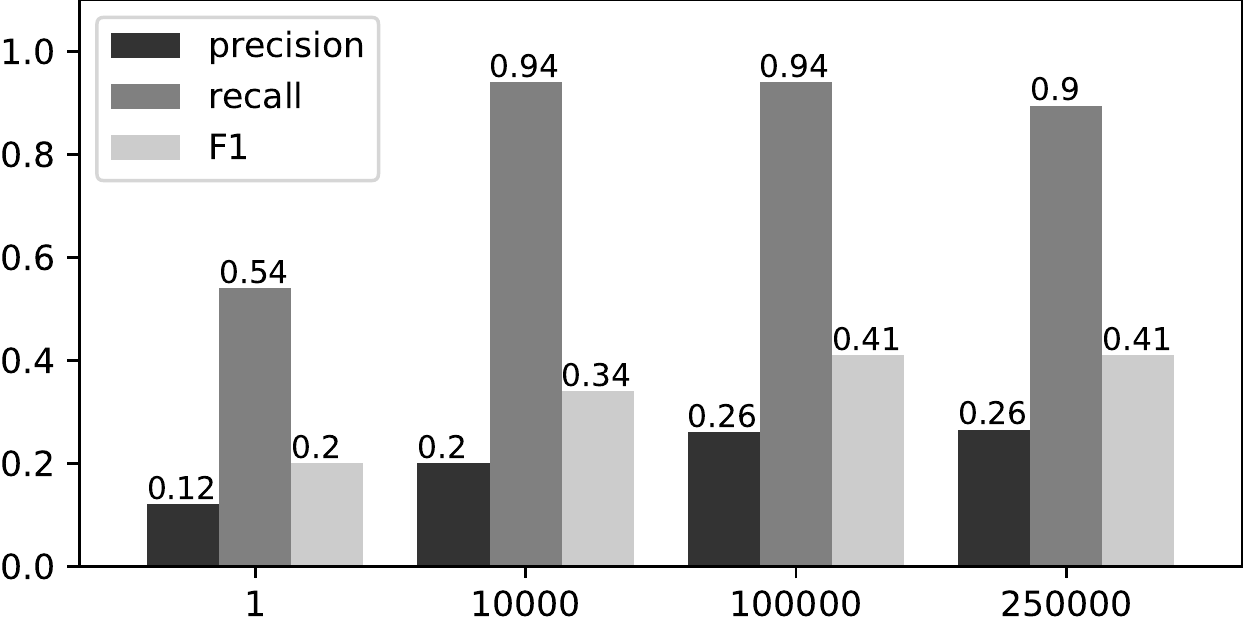}
\endminipage\hfill
\minipage{0.33\textwidth}
  \includegraphics[width=\linewidth]{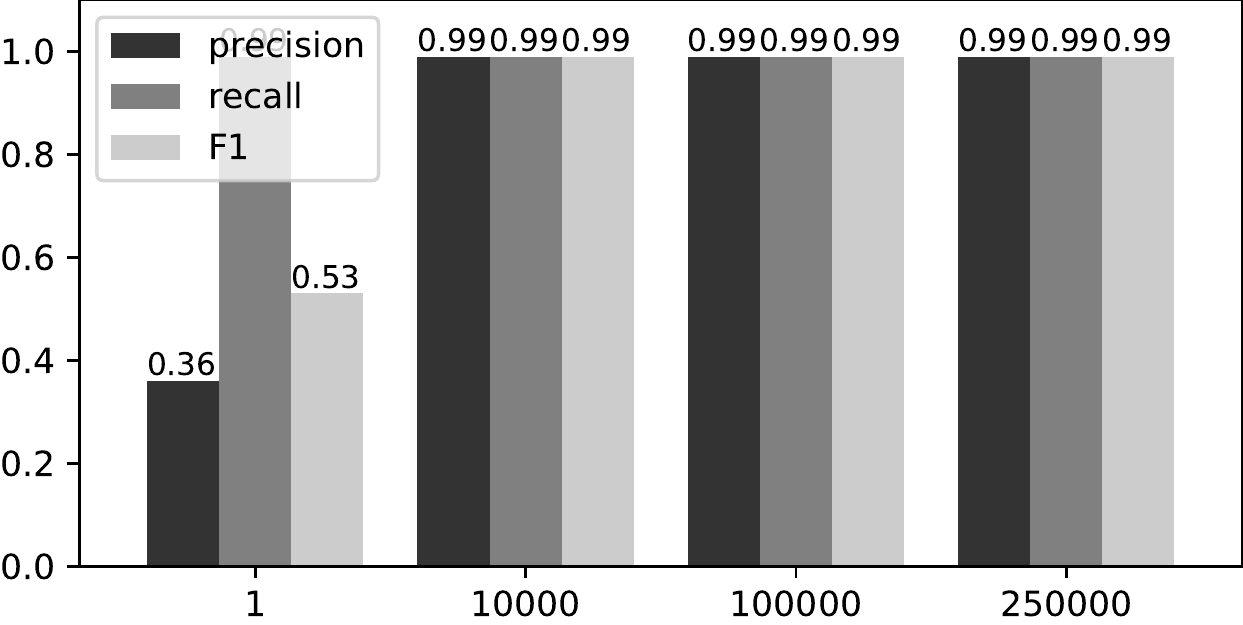}
\endminipage\hfill
\minipage{0.33\textwidth}
  \includegraphics[width=\linewidth]{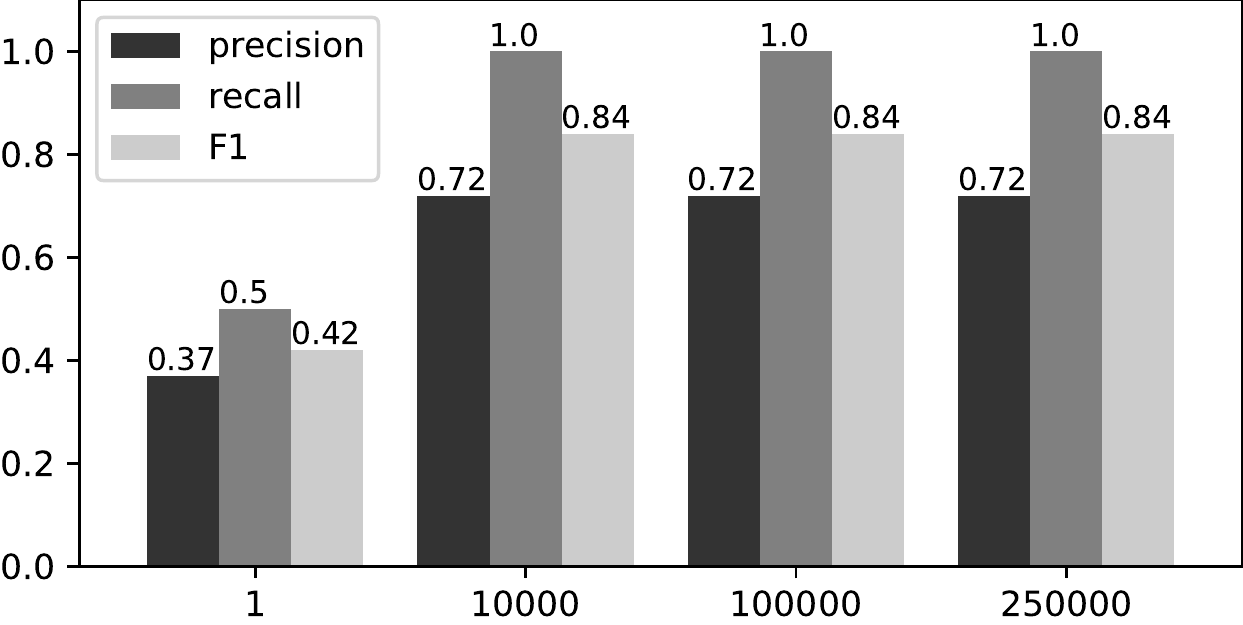}
\endminipage
\caption{Increasing the size auxiliary dataset, where the target system are Blue Gene/L, Thunderbird, and Spirit (left, middle, right) on 20\% train - 80\% test split}
\label{fig:auxiliary}
\end{figure*}

\begin{figure}[htbp]
    \centering
    \includegraphics[width=0.99\columnwidth]{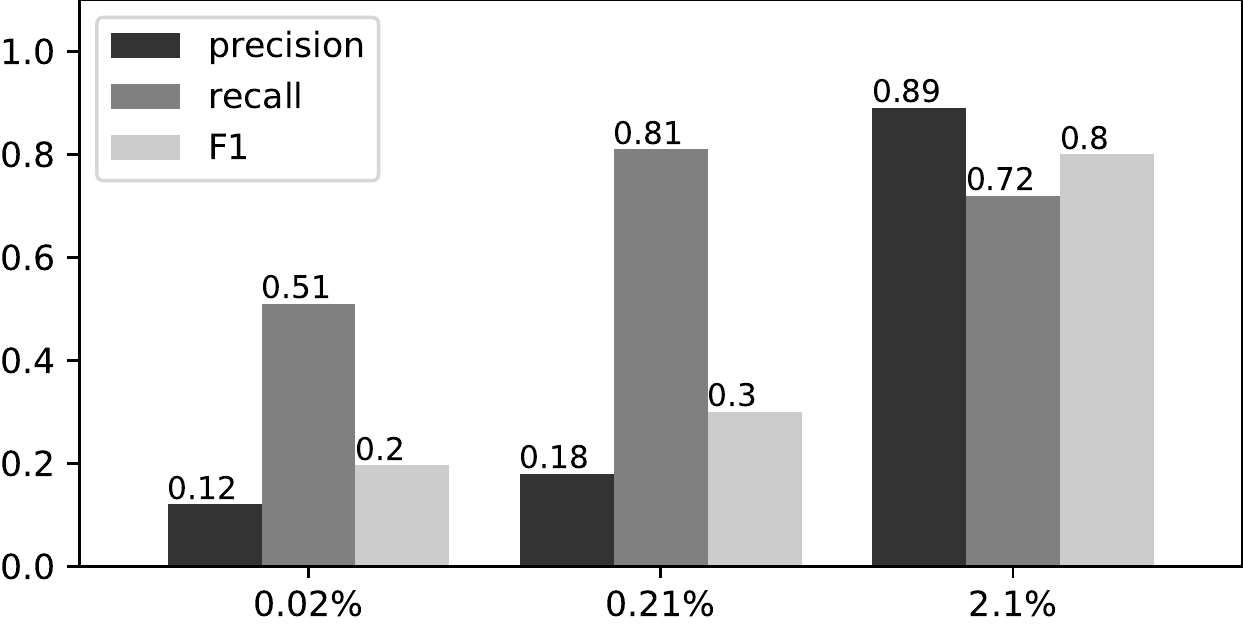}
    \caption{Increasing the size of the labeled anomaly data in the Blue Gene/L dataset (20\% train - 80\% test).}
    \label{fig:semisupervised}
\end{figure}

\subsubsection{The effect of the auxiliary data on the evaluation scores}
In this experiment, we perform an analysis of how Logsy performs when with various sizes of the auxiliary data. We evaluate the same target vs auxiliary data split for all datasets. We evaluate the approach on the 20\%-80\% train/test split. The results are shown in \figurename~\ref{fig:auxiliary} for all datasets. When the auxiliary data increases from 1 to 250000 we observe an increase in all evaluation scores. We observe that increasing the size of the auxiliary data from 100000 to 250000 the scores do not change in both cases. This shows that the amount of information present in the auxiliary data is similar and all cases are already present in 100000 random samples. We note that having just one auxiliary sample, which might even be generated artificially, sufficiently acts as a regularizer to the hypersphere loss function, preventing it from learning trivial solutions. Of course, increasing the variety of data (e.g., including more diverse log datasets) could further improve the performance, due to the increased number of samples representing abnormality.




\subsubsection{Including expert labeling}
Often systems are operated by a human operator which is an expert and has system-specific knowledge. Sometimes they could provide or manually label samples to improve the performance of the model. Here we experiment with the incremental inclusion of anomaly labels of the target dataset to test the model behaviour. We experiment on the 20\%-80\% split of the Blue Gene/L dataset. 
\figurename~\ref{fig:semisupervised} shows the results. Increasing the number of labelled anomaly samples improves performance. For as less as 2\% labelled data we already have the best performance of 0.8 F1-score. This shows that adding a few percentages of anomalies as labelled samples to Logsy, the performance dramatically improves. This only strengthens the hypothesis where the log anomaly detection must be addressed with an understanding of what causes a log message to be normal or anomaly. The labelled anomalies from the target system present information Logsy exploits to learn the differences between the normal and anomalous logs on the target dataset.

\begin{figure}[htbp]
    \centering
    \includegraphics[width=0.82\columnwidth]{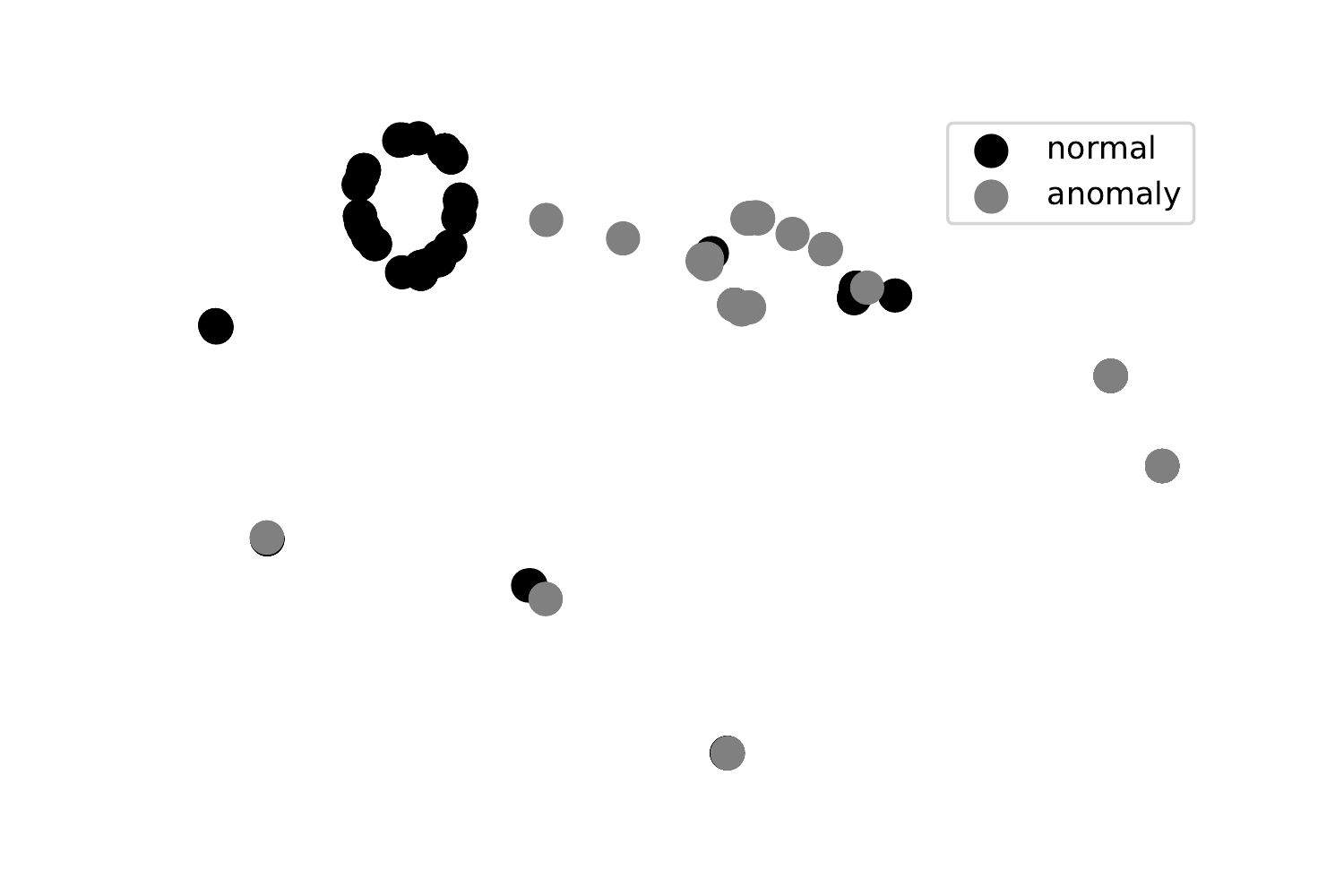}
    \caption{Visualisations of the log vector representations of Blue Gene/L with T-SNE~\cite{maaten2008visualizing}.}
    \label{fig:tsne}
\end{figure}

\subsubsection{Utilization of the learned log embeddings in related approaches}
In this experiment, we perform the extraction of the learned log message vector representations from the already trained Logsy. To illustrate the vector representations of the logs, in~\figurename~\ref{fig:tsne}, we show their lower-dimensional representation of the test split via the T-SNE dimensionality reduction method~\cite{maaten2008visualizing} on the Blue Gene/L dataset. We show that the log vector representations are somehow structured in a way following the definition of our spherical loss function (see Section~\ref{vectorrepresentations}). We can observe that the normal samples are concentrated around the centre of a hypersphere, which is a circle in two dimensions. Most of the anomalies are dispersed among the space outside of the sphere. Assigning a threshold on the anomaly score $A(x_i)$, i.e., the distance from the centre of the sphere (circle), we could obtain good performance. 

\begin{figure}[htbp]
    \centering
    \includegraphics[width=0.99\columnwidth]{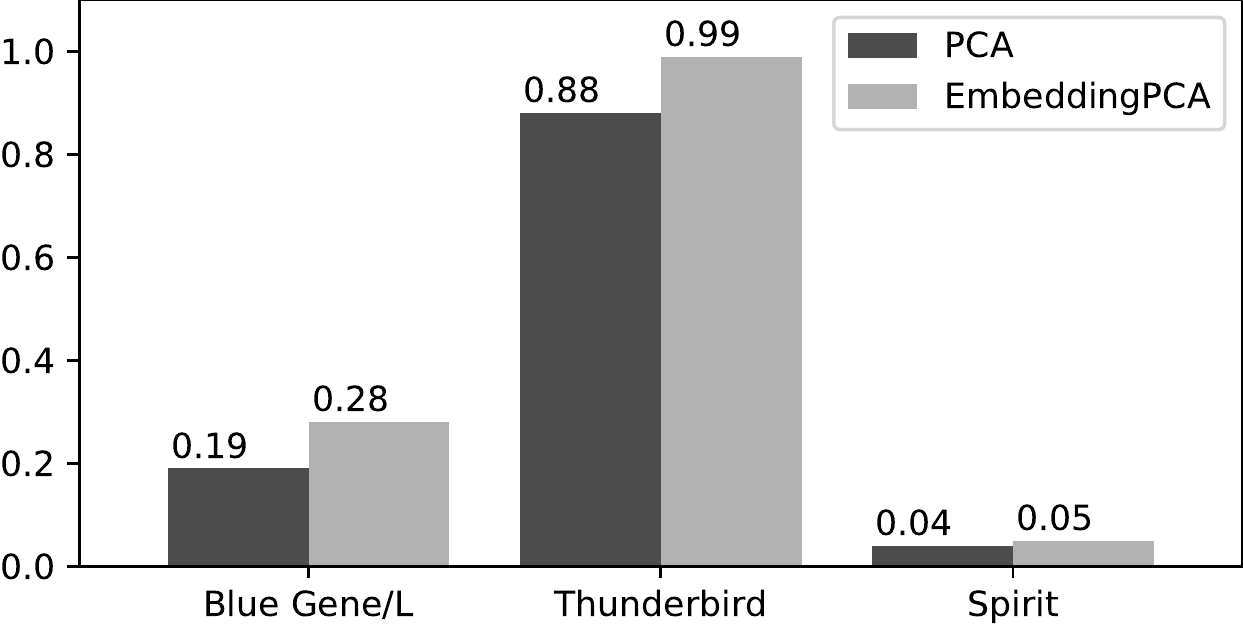}
    \caption{Comparison in F1 score between the standard PCA~\cite{xu2009detecting} and PCA using the embeddings extracted from our method (80\%-20\% split).}
    \label{fig:pcaembeddings}
\end{figure}

Furthermore, to evaluate the general importance of the log embeddings, we perform experiments where we replace the original TF-IDF log representations in PCA~\cite{xu2009detecting}, as the lowest-performing method, with the extracted embeddings from Logsy. We depict the results in the bar plot in \figurename~\ref{fig:pcaembeddings}. We observe that this replacement of the log representation improves the performance of PCA. We show improvement of 0.09, 0.11, and 0.01 F1-score for Blue Gene/L, Thunderbird, and Spirit respectively. This demonstrates that log representation learning has an impact, not only in Logsy, but also in previous approaches that could be adapted to use the new log embeddings. The relative improvement of the scores in average is 28.2\% in the F1-score.


\section{Conclusion}
Log anomaly detection is important to enhance the security and reliability of computer systems. Existing approaches lack generalization on new, unseen log samples, which comes from the evolution of logging statements as a consequence of system updates and the processing noise. To overcome this problem, we proposed a new anomaly detection approach, called Logsy. It is based on a self-attention encoder network with a hyperspherical classification objective. We formulated the log anomaly detection problem in a manner to discriminate between normal training data from the system of interest and samples from auxiliary easy-access log datasets from other systems, which represent an abnormality. We have presented experimental evidence that our classification-based method performs well for anomaly detection. The results of our method outperformed the baselines by a large margin of 0.25 F1 score. Furthermore, we demonstrated that the produced log vector representations could be utilized generally in other methods. We demonstrated that by adopting PCA to use the log vectors from Logsy, where we observed improvement of 0.07 (28.2\%) in the F1 score. 

We believe that future research on log anomaly detection should focus on finding alternative ways to incorporate richer domain bias emphasising the diversity of normal and anomaly data. 

\bibliographystyle{IEEEtran}
\bibliography{ICDM_submission}

\end{document}